\documentclass{sigplanconf}
\usepackage{graphicx}
\usepackage{amsmath}
\usepackage{amsfonts}
\usepackage{array}
\usepackage{textcomp}

\newcolumntype{L}{>{\centering\arraybackslash}m{3cm}}
\newcolumntype{D}{>{\centering\arraybackslash}m{1.7cm}}

\begin{document}

\title{Highly Efficient 8-bit Low Precision Inference of Convolutional Neural Networks with IntelCaffe}

\authorinfo{Jiong Gong, Haihao Shen, Guoming Zhang, Xiaoli Liu, Shane Li, Ge Jin, Niharika Maheshwari, Evarist Fomenko, Eden Segal}
{Intel Corporation}
{\{jiong.gong, haihao.shen, guoming.zhang, xiaoli.liu, li.shane, ge.jin, niharika.maheshwari, evarist.m.fomenko, eden.segal\}@intel.com}

\maketitle

\begin{abstract}
High throughput and low latency inference of deep neural networks are critical for the deployment of deep learning applications. This paper presents the efficient inference techniques of IntelCaffe, the first Intel\textsuperscript{\textregistered} optimized deep learning framework that supports efficient 8-bit low precision inference and model optimization techniques of convolutional neural networks on Intel\textsuperscript{\textregistered} Xeon\textsuperscript{\textregistered} Scalable Processors. The 8-bit optimized model is automatically generated with a calibration process from FP32 model without the need of fine-tuning or retraining. We show that the inference throughput and latency with ResNet-50, Inception-v3 and SSD are improved by 1.38X-2.9X and 1.35X-3X respectively with neglectable accuracy loss from IntelCaffe FP32 baseline and by 56X-75X and 26X-37X from BVLC Caffe. All these techniques have been open-sourced on IntelCaffe GitHub\footnote{https://github.com/intel/caffe}, and the artifact is provided to reproduce the result on Amazon AWS Cloud.
\end{abstract}

\section{Extended Abstract}

\subsection{Technical Description}
\begin{figure}
  \includegraphics[width=\linewidth]{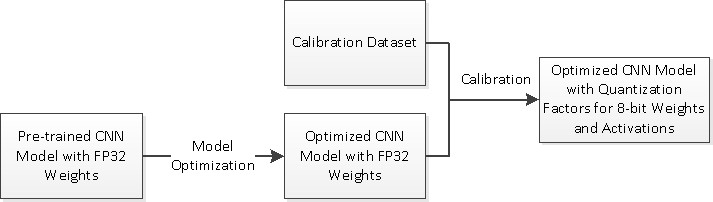}
  \caption{Steps IntelCaffe follows to generate the optimized CNN model that supports 8-bit low precision inference. The original FP32 model firstly goes through several model optimization steps (see \ref{sec:model_opt}). Then a calibration step generates quantization factors for each weight and activation according to the given dataset (see \ref{sec:int8}).}
  \label{fig:int8_workflow}
\end{figure}

While convolutional neural networks (CNN) shows state-of-the-art accuracy for wide range of computation vision tasks, it still faces challenges during industrial deployment due to its high computational complexity of inference. Low precision is one of the key techniques being actively studied recently to conquer the problem. With hardware acceleration support, low precision inference can compute more operations per second, reduce the memory access pressure and better utilize the cache, and deliver higher thoughput and lower latency. Intel\textsuperscript{\textregistered} Xeon\textsuperscript{\textregistered} Scalable Processors introduce the new instruction sets \cite{bib:int8_white_paper} to support efficient 8-bit low precision inference. Based on the hardware support and Intel\textsuperscript{\textregistered} Math Kernel Library for Deep Neural Networks (Intel\textsuperscript{\textregistered} MKL-DNN), IntelCaffe computes the CNN with 8-bit quantization of weights and activations. IntelCaffe also fuses several memory-bound operations and folds the learned parameters of batch normalization \cite{DBLP:journals/corr/IoffeS15} into convolution kernels to further shorten the inference execution time. Fig. \ref{fig:int8_workflow} shows the steps IntelCaffe follows to turn an FP32 CNN model into an optimized CNN model with quantization factors for 8-bit quantized form of FP32 weights and activations.

\subsubsection{8-bit Low Precision Inference}
\label{sec:int8}
In this sub-section, we first formulate the quantization method in IntelCaffe and then use it to explain the calibration and computation flow. More detailed explanation on the efficient 8-bit low precision algorithms that leverage hardware instructions can be found in \cite{bib:int8_white_paper}.
We define a quantization function $Q:\mathbb{R}^n\times \mathbb{R} \times \mathbb{N}\mapsto \mathbb{Z}^n\times \mathbb{R}$ in Eq.\ref{eq:Q} to turn an $n$-dimensional rational tensor ${\bf r}$ into an $n$-dimensional integer tensor ${\bf z}$ with the quantization factor $q$ and bit-precision $p$. Here $n$ could be of arbitrary dimensionality. The function $Round$ is a rounding function approximating a rational tensor with an integer tensor.
\begin{align} \label{eq:Q}
\begin{split}
  Q({\bf r},q,p) = Q_{p}({\bf r},q) = Q_{p,q}({\bf r}) = ({\bf z},q) = {\bf z}_q,
  \\
  {\bf z}=\max(\min(Round(\frac{{\bf r}}{q}),2^p-1),-2^p),\quad s.t.
  \\
  {\bf r}\in \mathbb{R}^n, q\in \mathbb{R}, p\in \mathbb{N}^+, {\bf z}\in \mathbb{Z}^n, Round:\mathbb{R}^n\mapsto \mathbb{Z}^n
\end{split}
\end{align}
We also define an inverse dequantization function $D:\mathbb{Z}^n\times \mathbb{R}\mapsto \mathbb{R}^n$ that approximates the rational tensor ${\bf r}$ with its quantized form ${\bf z}$ in Eq.\ref{eq:D}.
\begin{align} \label{eq:D}
  D({\bf z},q) = D_q({\bf z}) = D({\bf z}_q) = q{\bf z} = {\bf r'} \approx {\bf r}
\end{align}
We then define $+$ and $\times$ arithmetics on ${\bf z}_q$ in Eq.\ref{eq:arithmetics}. Here we assume $+$ and $\times$ have already been defined for tensor ${\bf r}$ and ${\bf z}$, e.g. when they are matrices.
\begin{align} \label{eq:arithmetics}
\begin{split}
  {\bf z}_{1,q_1} + {\bf z}_{2,q_2} = Q_p(D({\bf z}_{1,q_1})+D({\bf z}_{2,q_2}),\max(q_1,q_2))
  \\
  {\bf z}_{1,q_1}\times {\bf z}_{2,q_2} = ({\bf z}_1\times {\bf z}_2,q_1q_2)
\end{split}
\end{align}
IntelCaffe uses continuous quantization factor and symmetrical data range for minimal accuracy loss and efficient implementation. During the calibration flow, IntelCaffe samples each activation, weight and bias tensor on the given calibration dataset to get a maximum absolute value $max$ from each tensor and set the quantization factor of the tensor as $\frac{max}{2^p-1}$ where $p$ is the precision of quantization. $p=8$ is used for all non-negative activation tensors which are mostly true for popular CNN models after batch normalization operations are folded, ReLU non-linearities use zero negative slope \cite{DBLP:journals/corr/XuWCL15} and ReLU are fused into the convolution. For potentially negative activations, IntelCaffe simply falls back to FP32 since the hardware-accelerated 8-bit convolution only supports non-negative activations as input. $p=7$ is used for weight tensors and $p=31$ is used for bias. Then most activations and weights can be stored with 8-bit integer. IntelCaffe uses round-half-to-even as the $Round$ function for best statistical accuracy. Since both convolution and fully-connected computations can be modeled with affine transformation, we present the 8-bit low precision computation with the typical affine transformation with fused ReLU non-linearity ${\bf y}=ReLU({\bf Wx}+{\bf b})$ in Eq.\ref{eq:int8_affine} where ${\bf x}$ is the FP32 input, ${\bf y}$ the FP32 output, ${\bf W}$ the FP32 weight and ${\bf b}$ the FP32 bias. ${\bf y}_{q_y}$ is the quantized output. And similarly Eq.\ref{eq:int8_affine_fused} shows when the input tensor ${\bf z}$ is already in the quantized form from the previous computation, which is true in most situations.
\begin{align} \label{eq:int8_affine}
\begin{split}
  {\bf y}_{q_y} = Q_{8,q_y}({\bf y'})
  \\
  {\bf y'} = ReLU(D_{q_y}(Q_{7,q_W}({\bf W})\times Q_{8,q_x}({\bf x})+Q_{31,q_W+q_x}({\bf b})))
\end{split}
\end{align}
\begin{align} \label{eq:int8_affine_fused}
\begin{split}
  {\bf y}_{q_y} = Q_{8,q_y}({\bf y'})
  \\
  {\bf y'} = ReLU(D_{q_y}(Q_{7,q_W}({\bf W})\times {\bf z}_{q_z}+Q_{31,q_W+q_z}({\bf b})))
\end{split}
\end{align}
\subsubsection{Model Optimization}
\label{sec:model_opt}
\begin{figure}
  \includegraphics[width=\linewidth]{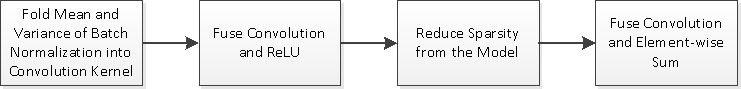}
  \caption{Four model optimization steps of IntelCaffe}
  \label{fig:model_optimization}
\end{figure}

Fig. \ref{fig:model_optimization} presents the model optimization pipeline of IntelCaffe. These optimization techniques can speed up both FP32 and 8-bit low precision models.

\begin{figure}
  \includegraphics[width=\linewidth]{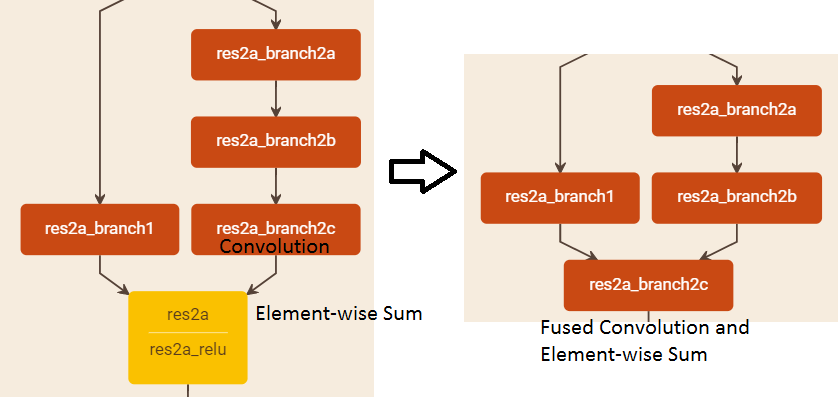}
  \caption{A typical residual block from ResNet model is shown. The convolution layer 'res2a\_branch2c' and element-wise sum layer 'res2a' on the left are fused into the new layer 'res2a\_branch2c' on the right. The ReLU layer 'res2a\_relu' is also fused into the convolution at the same time.}
  \label{fig:conv_sum_fusion}
\end{figure}

In most recent CNN models, a batch normalization operation is usually added after the convolution and in the inference computation, learned mean and variance are usually applied directly to the output activation of the convolution, which is equivalent to an affine transformation, therefore can be folded into the convolution kernel as follows. Both the new convolution weight $w'$ and bias $b'$ are affine transformation of the original weight $w$ and bias $b$. As was defined in \cite{DBLP:journals/corr/IoffeS15}, $\mu$ and $\sigma^{2}$ are the learned mini-batch mean and variance respectively, and $\gamma$ and $\beta$ are the scale and shift terms.
\begin{align*}
\begin{split}
 w'=(\frac{\gamma}{\sqrt{\sigma ^{2} +\epsilon }}) w ,
\\
 b'=(\frac{\gamma}{\sqrt{\sigma ^{2} +\epsilon }}) (b-\mu) + \beta .
\end{split}
\end{align*}
For modern CNN models, ReLU is the mostly used non-linearity placed after convolution. Intel\textsuperscript{\textregistered} MKL-DNN supports the fusion of convolution and ReLU to reduce the memory load and store operations from the standalone ReLU. Similar fusion technique is also implemented to reduce the memory load and store of element-wise sum operation at the end of a residual block where the convolution on one resdiual branch also does the element-wise sum over the result of the other branch before storing the result to memory (see Fig. \ref{fig:conv_sum_fusion}).

\begin{figure}
  \includegraphics[width=\linewidth]{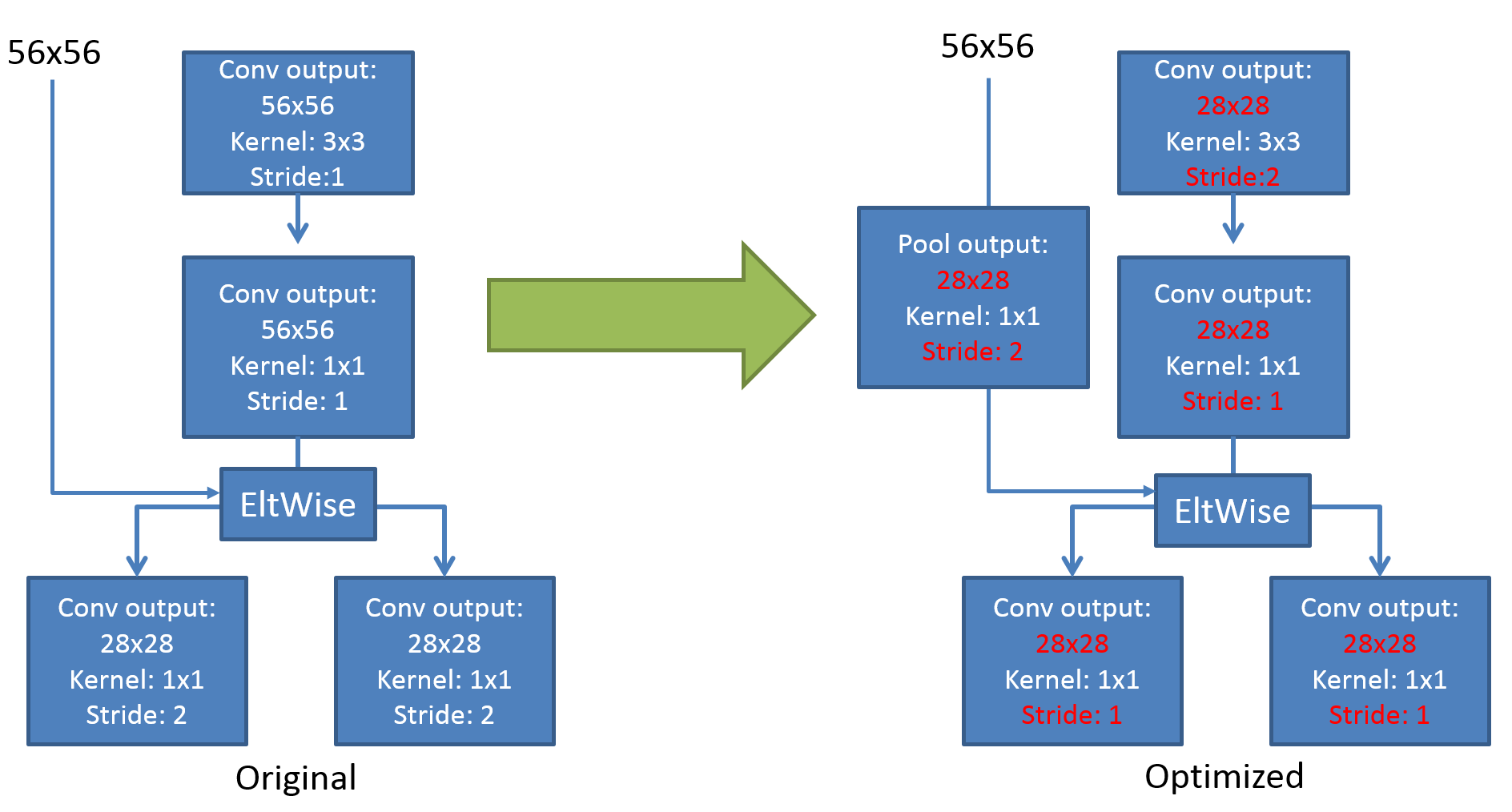}
  \caption{A typical residual structure from ResNet model is shown. Since the two 1x1 stride-2 convolution layers at the bottom only consume half of the activation, the optimization changes the in-bound layers to only produce the needed half data. 1x1 stride-2 pooling is added to the skip branch to match the other.}
  \label{fig:sparse_resnet}
\end{figure}

IntelCaffe also exploits the particular sparsity properties of ResNet family models and applies graph transformation to reduce the computation and memory access (see Fig. \ref{fig:sparse_resnet}).

\subsection{Empirical Evaluation}
We pick three popular CNN models ResNet-50 \cite{DBLP:journals/corr/HeZRS15}, Inception-v3 \cite{DBLP:journals/corr/SzegedyVISW15} and SSD with VGG-16 backbone \cite{DBLP:journals/corr/LiuAESR15}. We evaluated the optimization techniques on Amazon AWS Cloud using c5.18xlarge instance which has Intel\textsuperscript{\textregistered} Xeon\textsuperscript{\textregistered} Platinum 8124M Processor. The optimization techniques introduced in this paper bring additional 1.38X-2.9X and 1.35X-3X over the FP32 baseline. The overall throughput speedup and latency reduction of IntelCaffe over BVLC Caffe are 56X-75X and 26X-37X due to IntelCaffe optimization and the usage of Intel\textsuperscript{\textregistered} MKL-DNN (see Table \ref{tab:baremetal-throughput} and \ref{tab:baremetal-latency}). The data of SSD on BVLC Caffe is not available due to lack of support.

We also compare the statistical accuracy of 8-bit low precision inference on ImageNet-1k \cite{ILSVRC15} and PASCAL VOC2007 \cite{Everingham15} datasets with FP32 baseline in Table \ref{tab:accuracy}. The accuracy loss is all less 1\% point.

\begin{table}[]
\centering
\small
\setlength\tabcolsep{2pt}
\caption{Inference throughput (images/second) of popular CNN models after adding optimization techniques in IntelCaffe, measured on single-socket of c5.18xlarge.}
\label{tab:baremetal-throughput}
\begin{tabular}{LDDD}
\hline
           & ResNet-50 & Inception-v3 & SSD \\
\hline
BVLC Caffe &  6.1         &     5.0         &    N/A \\
\hline
IntelCaffe FP32 Baseline & 158 & 126 & 31 \\
\hline
+Folded BatchNorm & 189 & 144 & 31 \\
\hline
+Fused Convolution and ReLU & 199 & 175 & 34 \\
\hline
+Remove Sparsity & 225 & 175 & 34 \\
\hline
+Fused Convolution and Element-wise Sum & 312 & 175 & 34 \\
\hline
+8-bit Low Precision & \textbf{462} & \textbf{282} & \textbf{43} \\
\hline
\end{tabular}
\end{table}

\begin{table}[]
\centering
\small
\setlength\tabcolsep{2pt}
\caption{Inference latency (milliseconds) of popular CNN models after adding optimization techniques in IntelCaffe, measured on single-socket of c5.18xlarge.}
\label{tab:baremetal-latency}
\begin{tabular}{LDDD}
\hline
           & ResNet-50 & Inception-v3 & SSD \\
\hline
BVLC Caffe &    131.8       &    158.96          &  N/A   \\
\hline
IntelCaffe FP32 Baseline & 10.7 & 14.3 & 33.7 \\
\hline
+Folded BatchNorm & 7.8 & 10.6 & 33.7 \\
\hline
+Fused Convolution and ReLU & 7.5 & 9.5 & 32.9 \\
\hline
+Remove Sparsity & 7.0 & 9.5 & 32.9 \\
\hline
+Fused Convolution and Element-wise Sum & 6.2 & 9.5 & 32.9 \\
\hline
+8-bit Low Precision & \textbf{3.5} & \textbf{5.9} & \textbf{24.8}\\
\hline
\end{tabular}
\end{table}

\begin{table}[]
\centering
\small
\setlength\tabcolsep{2pt}
\caption{Statistical accuracy (\%) of 8-bit low precision inference on 3 popular CNN models, ImageNet-1k used for ResNet-50 and Inception-v3, PASCAL VOC2007 for SSD.}
\label{tab:accuracy}
\begin{tabular}{LDDD}
\hline
           & ResNet-50 (top-1/top-5) & Inception-v3 (top-1/top-5) & SSD (mAP) \\
\hline
FP32 Baseline & 72.5/90.87 & 75.44/92.36 & 77.68 \\
\hline
8-bit Low Precision & 71.84/90.49 & 75.31/92.33 & 77.51 \\
\hline
\end{tabular}
\end{table}

\bibliographystyle{plainnat}
\bibliography{main}
\newpage

\appendix
\section{Artifact appendix}

\subsection{Abstract}

Hardware and software configurations and execution steps are provided to reproduce the inference throughput and latency result mentioned in this paper.

\subsection{Artifact check-list (meta-information)}

{\small
\begin{itemize}
  \item {\bf Compilation: } Intel C++ Compiler 17.0.5 20170817
  \item {\bf Binary: } \$CAFFE\_ROOT/build/tools/caffe
  \item {\bf Data set: } Fake random data for throughput and latency. ImageNet-1k and PASCAL VOC2007 for accuracy.
  \item {\bf Run-time environment: }
  KMP\_HW\_SUBSET=1T \\
  KMP\_AFFINITY=granularity=fine,compact \\
  OMP\_NUM\_THREADS=18
  \item {\bf Hardware: } single socket (18 cores) on c5.18xlarge.
  \item {\bf Execution: } python benchmark.py -mode accuracy[fps$|$throughput]
  \item {\bf Metrics: } \\ Throughput: images per second. \\ Latency: milli-second. \\ Accuracy: \% top-1/top-5/mAP.
  \item {\bf Output: } Standard console output (stdout)
  \item {\bf Experiments: } We use batch size 64, 64, and 32 to measure the throughput for ResNet-50, Inception-V3, and SSD respectively. We use batch size 1 to measure the latency.
  \item {\bf Workflow frameworks used?: } IntelCaffe and BVLC Caffe\footnote{https://github.com/BVLC/caffe.git}
  \item {\bf Publicly available?: } Yes, available on GitHub with wiki installation guide.
\end{itemize}

\begin{itemize}
  \item {\bf Artifacts publicly available?: } Yes.
  \item {\bf Artifacts functional?:} Yes.
  \item {\bf Artifacts reusable?:} Yes.
  \item {\bf Results validated?:} Yes.
\end{itemize}

\subsection{Description}

\subsubsection{How delivered}
AWS Public Deep Learning Image containing pre-built BVLC Caffe, IntelCaffe GitHub.
\subsubsection{Hardware dependencies}
Amazon AWS c5.18xlarge instance.
\subsubsection{Software dependencies}
IntelCaffe (branch: request\_artifact), BVLC Caffe, Intel\textsuperscript{\textregistered} C++ Compiler
\subsubsection{Data sets}
Fake dataset, ImageNet-1k and PASCAL VOC2007
\subsection{Installation}
https://github.com/intel/caffe/wiki/ReQuEST-Artifact-Installation-Guide

\subsection{Experiment workflow}
On throughput measurement, we use batch size 64, 64, and 32 for ResNet-50, Inception-V3, and SSD, respectively. We measure the latency with the single batch on one socket. On accuracy, the auxiliary script would first calibrate the FP32 models of ResNet-50, Inception-V3 and SSD on the ImageNet-1k and PASCAL VOC2007 dataset, generating 8-bit low precision prototxt files. Then these files can be executed on the same datasets to get the test accuracy.
\subsection{Evaluation and expected result}
You will get the similar performance data within 5\% range for both throughput and latency. Since AWS instances are virtualized, there might be larger variances than bare metal machines. I recommend the evaluators to collect the performance data three times to pick the stable data points. On accuracy, you will get equivalent accuracy result.

\subsection{Experiment customization}

\subsection{Notes}
For SSD, we added DetectOutput layer in order to calculate the mAP.

\end{document}